\documentclass[runningheads]{llncs}
\usepackage{graphicx}
\usepackage{amsmath,amssymb} 
\usepackage{color}

\usepackage{epstopdf}
\usepackage{comment}
\usepackage{booktabs}
\usepackage{caption}
\usepackage{subcaption}
\usepackage[colorlinks=true,
           citecolor=red,
            linkcolor=blue,
           anchorcolor=green,
            urlcolor=red]{hyperref}

\begin{document}
\newcommand{\point}{
    \raise0.7ex\hbox{.}
    }
\pagestyle{headings}

\mainmatter

\title{Object-Centric Representation Learning \\from Unlabeled Videos} 

\titlerunning{Object-Centric Representation Learning \\from Unlabeled Videos} 

\authorrunning{Ruohan Gao, Dinesh Jayaraman, Kristen Grauman} 

\author{Ruohan Gao, Dinesh Jayaraman, Kristen Grauman} 
\institute{University of Texas at Austin} 

\maketitle

\begin{abstract}
Supervised (pre-)training currently yields state-of-the-art performance for representation learning for visual recognition, yet it comes at the cost of (1) intensive manual annotations and (2) an inherent restriction in the scope of data relevant for learning.  In this work, we explore unsupervised feature learning from unlabeled video.  We introduce a novel \emph{object-centric} approach to temporal coherence that encourages similar representations to be learned for object-like regions segmented from nearby frames.  Our framework relies on a Siamese-triplet network to train a deep convolutional neural network (CNN) representation.  Compared to existing temporal coherence methods, our idea has the advantage of lightweight preprocessing of the unlabeled video (no tracking required) while still being able to extract object-level regions from which to learn invariances. Furthermore, as we show in results on several standard datasets, our method typically achieves substantial accuracy gains over competing unsupervised methods for image classification and retrieval tasks.
\end{abstract}

\vspace{-3pt}
\section{Introduction}
The emergence of large-scale datasets of millions of labeled examples such as ImageNet has led to major successes for supervised visual representation learning. Indeed, visual feature learning with deep neural networks has swept the field of computer vision in recent years \cite{krizhevsky2012imagenet,simonyan2014very,he2015deep}. If learned from labeled data with broad enough coverage, these learnt features can even be transferred or repurposed to other domains or new tasks via ``pretraining" \cite{donahue2013decaf,girshick2014rich}. However, all such methods heavily rely on ample manually provided image labels. Despite advances in crowdsourcing, massive labeled image datasets remain quite expensive to collect.  Furthermore, even putting cost aside, it is likely that restricting visual representation learning to a bag of unrelated images (like web photos) may prevent algorithms from learning critical properties that simply are not observable in such data---such as certain invariances, dynamics, or patterns rarely photographed in web images.

\begin{figure}[ht]
      \centering
      \includegraphics[scale=0.2]{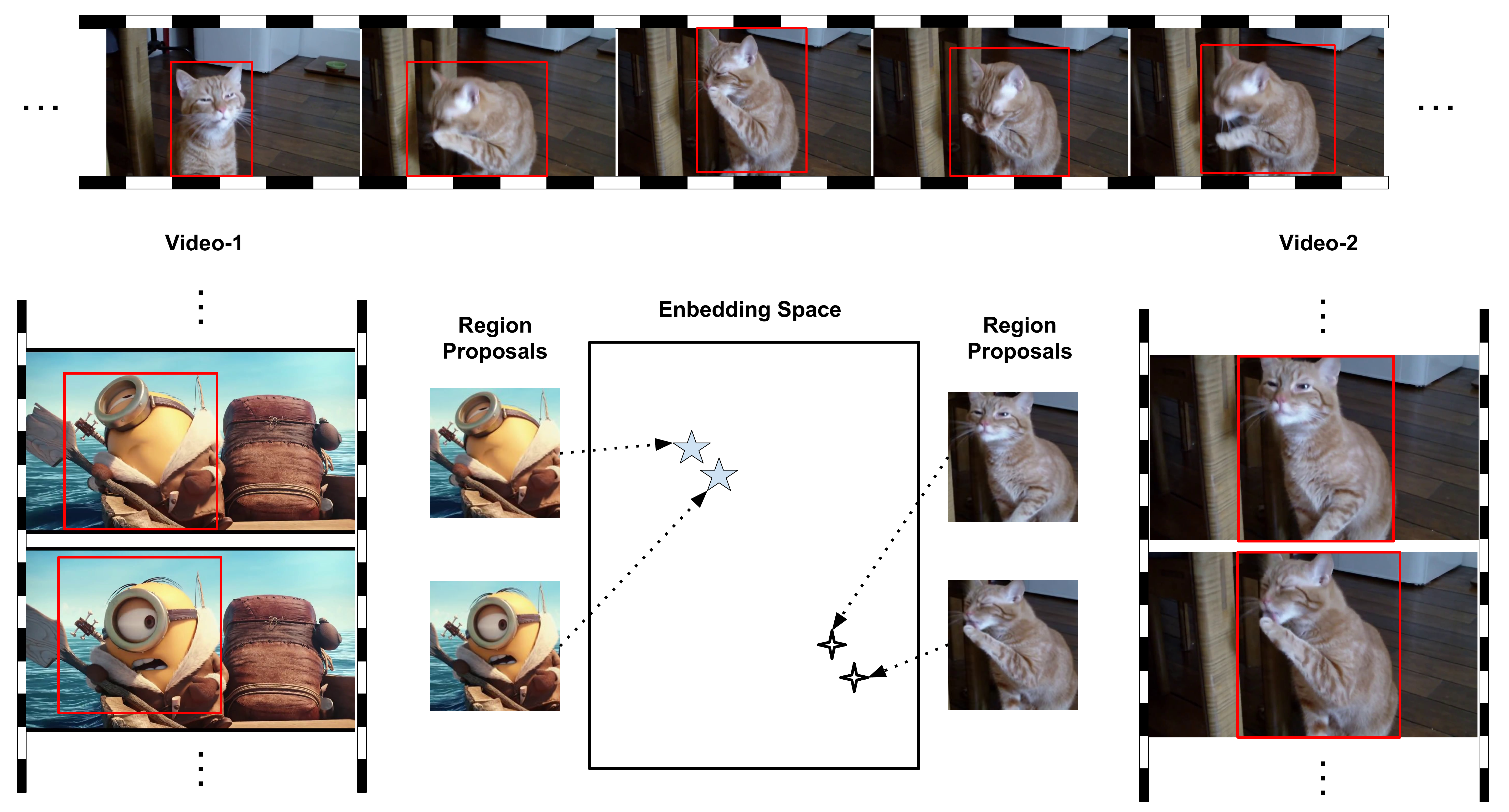}
            \caption{Video frames tend to change coherently across the video, as shown in the two example videos above. These changes usually come from certain objects in the frames. In our framework, we first generate region proposals independently on whole video frames and find spatio-temporally close region proposal pairs. Instead of embedding video frames directly, we embed these region proposals. Spatio-temporally close region proposal pairs should be close in the deep embedding space, since they are likely to belong to the same object or object part.}
            \vspace{-1.5em}
            \label{fig:framework}
\end{figure}

Due to these restrictions, \emph{unsupervised visual feature learning} from unlabeled images or videos has therefore increasingly drawn  researchers' attention.  An unsupervised approach has several potential advantages.  First, it is in principle much more scalable, because unlabeled images and videos can be obtained essentially for free.  Moreover, features learnt in an unsupervised way, particularly from diverse videos, may prove even more effective as a  generalizable base representation by being unencumbered by the ``closed world" restriction of categorically labeled data.  Recent years have seen a number of exciting ideas in unsupervised visual feature learning, particularly using temporal coherence  of unlabeled video~\cite{mobahi2009deep,wang2015unsupervised,ramanathan2015learning,goroshin2015unsupervised,jayaraman2015slow}, self-supervision from image context~\cite{doersch2015unsupervised,pathak2016context} or ego-motion~\cite{agrawal2015learning,jayaraman2015learning}, as well as earlier attempts based on autoencoders~\cite{le2012building,bengio2007greedy,bengio2009learning,hinton2006reducing}.

In this work, we focus on learning from unlabeled videos and build upon the idea of \emph{temporal coherence} as a form of ``free" supervision to learn image representations invariant to small transformations. Temporal coherence, a form of \emph{slow feature analysis}~\cite{wiskott2002slow}, is based on the observation that high-level signals cannot change too quickly from frame to frame; therefore, temporally close frames should also be close in the learned feature space.  Most prior work in this space produces a holistic image embedding, and attempts to learn feature representations for video frames as a whole \cite{mobahi2009deep,bengio2009slow,goroshin2015unsupervised,ramanathan2015learning,jayaraman2015slow}. Two temporally close video frames, although similar, usually have multiple layers of changes across different regions of the frames. This may confuse the deep neural network, which tries to learn good feature representations and embed these two frames in the deep feature space as a whole. An alternative is to track local patches and learn a localized representation based on the tracked patches~\cite{wang2015unsupervised,zou2012deep,zou2011unsupervised}. In particular, a recent approach~\cite{wang2015unsupervised} uses sophisticated visual motion tracking to connect ``start" and ``end" patches for training pairs. However, such tracking is biased towards moving objects, which may limit the invariances that are possible to learn. Furthermore, processing massive unlabeled video collections with tracking algorithms is computationally intensive, and errors in tracking may influence the quality of the patches used for learning.

With these limitations in mind, we propose a new way to learn visual features from unlabeled video.  Similar to existing methods, we exploit the general principle of temporal coherence.  Unlike existing methods, however, we neither learn from whole-frames of video nor rely on tracking fragments in the video.  Instead, we propose to focus temporal coherence on \emph{object-centric} regions discovered with object proposal regions. In particular, we first generate object-like region proposals on temporally adjacent video frames using Selective Search~\cite{uijlings2013selective}. Then we perform feature learning using a ranking-based objective that maps spatio-temporally adjacent region proposals closer in the embedding space than non-neighbors.  The idea is that two spatio-temporally close region proposals should be embedded close in the deep feature space since they likely belong to the same object or object part, in spite of superficial differences in their pose, lighting, or appearance.  See Fig.~\ref{fig:framework}.

Why might such an object-centric approach have an advantage?  How might it produce features better equipped for object recognition, image classification, or related tasks? First, unlike patches found with tracking, patches generated by region proposals can capture static objects as well as moving objects. Static objects are also informative in the sense that, beyond object motion, there might be other slight changes such as illumination changes or camera viewpoint changes across video frames. Therefore, static objects should not be neglected in the learning process. Secondly, our framework can also help capture the object-level regions of interest in cases where there is motion only on a part of the object.  For example, in Fig.~\ref{fig:framework}, the cat is moving its paw but otherwise staying similarly posed. Visual motion tracking will have difficulty catching such subtle changes (or will catch only the small moving paw), while region proposals used as we propose can easily capture the entire object with its part-level change. Thirdly, our method is much more efficient to generate training samples since no tracking is needed.

In results on three challenging datasets, we show the impact of our approach for unsupervised convolutional neural network (CNN) feature learning from video.  Compared to the alternative whole-frame and tracking-based paradigms discussed above, our idea shows consistent advantages.  Furthermore, it often outperforms an array of state-of-the-art unsupervised feature learning methods~\cite{pathak2016context,agrawal2015learning,wang2015unsupervised} for image classification and retrieval tasks.  In particular, we observe relative gains of 10 to 30\% in most cases compared to existing pre-trained models. Overall, our simple but effective approach is an encouraging new path to explore learning from unlabeled video.

\vspace{-3pt}

\section{Related Work}
\label{related_work}
Unsupervised feature learning has a rich history and can be traced to seminal work for learning visual representations which are sparse and reconstructive~\cite{olshausen1997sparse}.  More recent advances include training a deep belief network  by stacking layer-by-layer RBMs~\cite{hinton2006reducing} and injecting autoencoders~\cite{bengio2007greedy}. Building on this concept, multi-layer autoencoders are scaled up to large-scale unlabeled data~\cite{le2012building}, where it is shown that neurons in high layers of an unsupervised network can have high responses on semantic objects or object parts. Recently, some approaches explore the use of spatial context of images as a source of a (self-)supervisory signal for learning visual representations~\cite{doersch2015unsupervised,pathak2016context}. In \cite{doersch2015unsupervised}, the learning is driven by position prediction of context patches, while in \cite{pathak2016context}, the algorithm is driven by context-based pixel prediction in images.

Most existing work for learning representations from unlabeled video exploits the concept of \emph{temporal coherence}.  The underlying idea can be traced to the concept of slow feature analysis (SFA) \cite{wiskott2002slow,hurri2003simple}, which proposes to use temporal coherence in a sequential signal as ``free'' supervision as discussed above.  Some methods attempt to learn feature representations of video frames as a whole \cite{mobahi2009deep,bengio2009slow,goroshin2015unsupervised,jayaraman2015slow}, while others track local patches to learn a localized representation \cite{wang2015unsupervised,zou2012deep,zou2011unsupervised}. Our approach builds on the concept of temporal coherence, with the new twist of learning from localized object-centric regions in video, and without requiring tracking. 

Another way to learn a feature embedding from video is by means of a ``proxy" task, solving which entails learning a good feature embedding. For example, the reconstruction and prediction of a sequence of video frames can serve as the proxy task~\cite{srivastava2015unsupervised,ramanathan2015learning}. The idea is that in order to reconstruct past video frames or predict future frames, good feature representations must be learnt along the way. Ego-motion \cite{agrawal2015learning,jayaraman2015learning} is another interesting proxy that is recently adopted to learn feature embeddings. Learning the type of ego-motion that corresponds to video frame transformations entails learning good visual features, and thus proprioceptive motor signals can also act as a supervisory signal for feature learning.  We offer empirical comparisons to recent such methods, and show our method surpasses them on three challenging datasets.

\vspace{-3pt}

\section{Our Framework}
Given a large collection of unlabeled videos, our goal is to learn image representations that are useful for generic recognition tasks. Specifically, we focus on learning representations for \emph{object-like} regions. Our key idea is to learn a feature space where representations of object-like regions in video vary slowly over time. Intuitively, this property induces invariances (such as to pose, lighting etc.) in the feature space that are useful for high-level tasks. Towards this goal, we start by generating object-like region proposals independently on each frame of hundreds of thousands of unlabeled web videos (details below). We observe that objects in images vary slowly over time, so that correct region proposals in one frame tend to have corresponding proposals in adjacent frames (1s apart in our setting). Conversely, region proposals that are spatio-temporally adjacent usually correspond to the same object or object part. For example, in Fig.~\ref{fig:framework}, semantically meaningful objects like the cartoon character minions and the cat are proposed in the video frames by Selective Search. Based on this observation, we train deep neural networks to learn feature spaces where spatio-temporally adjacent region proposals in video are embedded close to each other. Next, we describe these two stages of our approach in detail: (1) region proposal pair selection (Sec.~\ref{region_proposal}), and (2) slow representation learning (Sec.~\ref{sec:slow_learn}).

\vspace{-6pt}

\subsection{Selecting Region Proposal Pairs from Videos}
\label{region_proposal}

\vspace{-8pt}

We start by downloading hundreds of thousands of YouTube videos (details in Sec.~\ref{training_procedures}). Among these, we are interested in mining region proposal pairs that (1) correspond to the same object and (2) encode useful invariances.

\vspace{-6pt}

\subsubsection{Whole Frame Pair Selection}
We start by selecting frame pairs likely to yield useful region proposals, based on two factors:

\vspace{-8pt}

\paragraph{\textbf{Pixel Correlation}}
We extract video frames at a frame rate of fps = 1, and every two adjacent frames form a candidate video frame pair. We compute pixel space correlations for all frame pairs. Very low correlations usually correspond to scene cuts, and very high correlations to near-static scenes, where the object-like regions do not change in appearance, and are therefore trivially mapped to the same feature representation. Thus, neither of these cases yields region pairs useful for slow representation learning. Therefore, at this stage, we only select frame pairs whose correlation score $\in(0.3,0.8)$. 

\vspace{-8pt}

\paragraph{\textbf{Mean Intensity}}
 Next, we discard video frame pairs where either one of the frames has a mean intensity value lower than 50 or larger than 200. These are often ``junk'' frames which do not contain meaningful region proposals, such as the prologue or epilogue of a movie trailer.

\begin{figure}
      \centering
      \vspace{-1em}
      \includegraphics[scale=0.36]{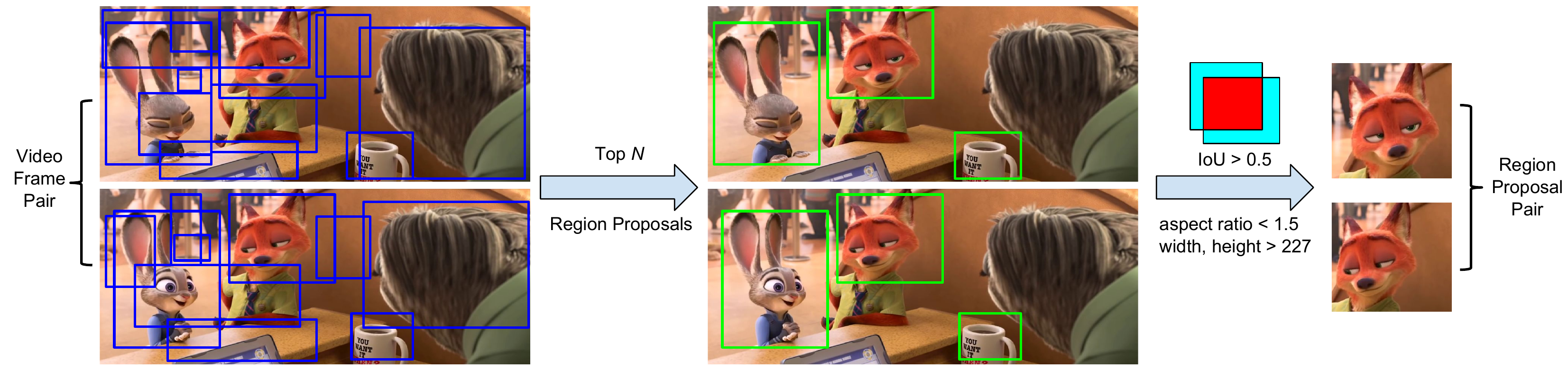}
            \caption{A large quantity of region proposals can be generated in each video frame. We only keep the top-N scored region proposals (N = 100 in our experiments, N = 3 in the figure above for demonstration). For one region proposal in the first frame, we find the region proposal in the other frame with the largest IoU score. If the two spatio-temporally adjacent region proposals satisfy the thresholds on IoU, width, height and aspect ratio, two $227 \times 227$ regions are then cropped from them to form a region proposal pair. }
            \vspace{-1em}
            \label{fig:filter_process}
\end{figure}

\vspace{-20pt}

\subsubsection{Proposal Pair Selection}
\label{region_proposal}
We now generate region proposal pairs from these selected video frame pairs using a standard object propsal generation method, as illustrated in Fig.~\ref{fig:filter_process}. We use Selective Search~\cite{uijlings2013selective}, which generates hundreds of region proposals from each frame. Starting from the large candidate pool of all region proposal pairs from adjacent frames, we use the following filtering process to guarantee the \emph{quality}, \emph{congruity}, and \emph{diversity} of our region proposal pairs:

\paragraph{\textbf{Quality}}
We only keep the top-100 scored region proposals from Selective Search for each frame. These region proposals are of higher quality and tend to be more object-like. Furthermore, we only keep region proposals of width and height both larger than 227 and aspect-ratio smaller than 1.5. We then crop $227 \times 227$ regions to get the final region proposal pairs.

\vspace{-6pt}

\paragraph{\textbf{Congruity}}
As mentioned above, regions corresponding to the same object are likely to be spatially close in adjacent frames, and conversely, spatially close high quality proposals in adjacent frames usually correspond to the same object. With this in mind, at this stage, we only retain those region proposals from neighboring frames that have a spatial overlap score (intersection over union) exceeding 0.5. Where there are multiple candidate pairings for a single region proposal, we only retain the pairing with the largest IoU score.

\vspace{-6pt}

\paragraph{\textbf{Diversity}}
To increase the diversity of our region proposal pairs and avoid redundant pairs, we process each video sequentially, and compute the pixel-space correlation of each candidate pair with the last selected pair from the same video. We save the current pair only if this correlation with the last selected pair is $<0.7$. For computational efficiency, we first downsample the region patches to $33 \times 33$ and then calculate the correlation. This step reduces redundancy and ensures that each selected pair is more likely to be informative.

\begin{figure}
      \centering
      \includegraphics[scale=0.31]{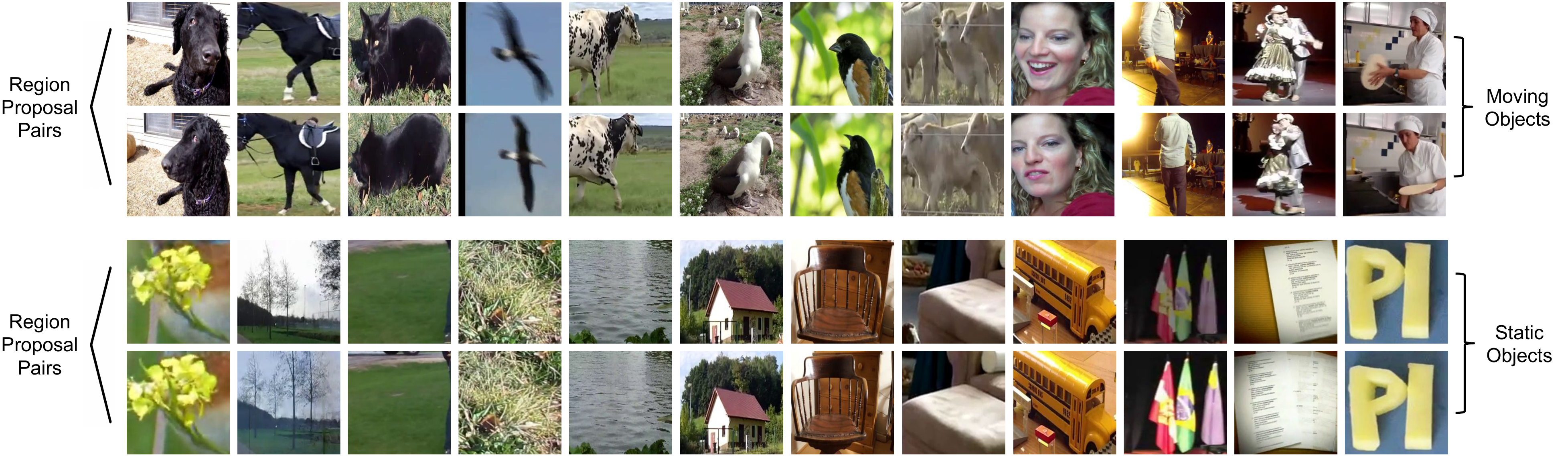}
            \caption{Examples of region proposal pairs extracted from unlabeled videos. Note that both moving objects (e.g. dog, horse, cat, human, etc) and static objects (e.g. flower, tree, chair, sofa, etc) can be captured by region proposals. Moving objects or object parts (like moving animals and human beings) are informative. Changes like illumination variations and camera viewpoint changes across static objects (from both natural scenes and indoor scenes) also form informative pairs from which to learn.}
            \vspace{-1.5em}
            \label{fig:region_proposal_example}
\end{figure}

\noindent
Our method of generating patch pairs is much more efficient than tracking \cite{wang2015unsupervised}. Our lightweight approach takes only a minute to generate hundreds of region proposal pairs, which is more than 100 times faster than tracking (our implementation). Fig.~\ref{fig:region_proposal_example} shows some examples of the region proposal pairs selected by our approach. We can see that these regions tend to belong to objects or object parts. Region pairs corresponding to both static and moving object regions are selected. The difference between the two region proposals in a pair comes from various sources: movements of the captured objects, partial movement of object-level regions, illumination changes across frames, viewpoint changes, etc. A full coverage of different types of changes among region proposal pairs can help to guarantee useful invariances in features trained in the next stage.

\vspace{-1em}
\subsection{Learning from Region Proposal Pairs}
\label{sec:slow_learn}
After generating millions of region proposal pairs from those unlabeled videos, we then train a convolutional neural network based on these generated pairs. The trained CNN is expected to map region proposal pairs from image space to the feature space. Specifically, given a region proposal $R$ as an input for the network, we aim to learn a feature mapping $f: R \Rightarrow f(R)$ in the final layer. Let $(R_1$, $R_2)$ denote a region proposal pair (spatio-temporally close region proposals) from one video, and $R^{-}$ denote a randomly sampled region proposal from another video\footnote{25,000 videos are used to generate training samples. The chance that the object proposal from one video and a random proposal from another video are similar (or of the exact same object instance) is negligible.}. In our target feature space, $R_1$ and $R_2$ should be closer to each other compared to $R^{-}$, i.e.~$D(R_1,R^{-}) > D(R_1,R_2)$, where $D(\cdot)$ is the cosine distance in the embedding space:

\vspace{-2pt}
			$$D(R_1, R_2) = 1 - \frac{f(R_1)\cdot f(R_2)}{\parallel f(R_1) \parallel \parallel f(R_2)\parallel}.$$
\vspace{-1pt}

\begin{figure}[ht]
      \centering
      \vspace{-1.5em}
      \includegraphics[scale=0.4]{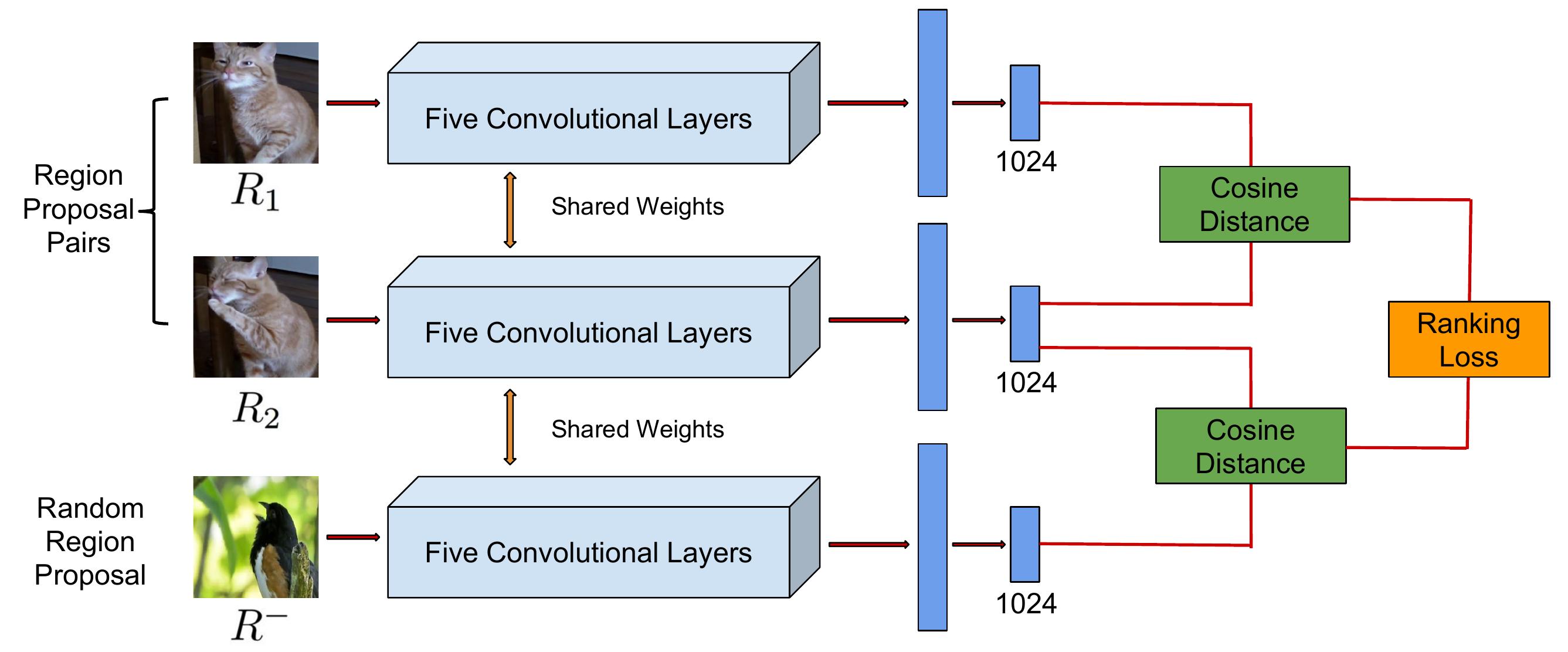}
            \caption{The Siamese-triplet network consists of three base networks with shared parameters. The five convolutional layers of the base network are the same as the architecture of AlexNet, and two fully connected layers are stacked on the pool5 outputs. The final ranking loss is defined based on the last fully connected layers. The whole Siamese-triplet network is trained from scratch, using only our region proposals as input data.}
            \vspace{-1.5em}
            \label{fig:siamese}
\end{figure}

We use a ``Siamese-triplet'' network to learn our feature space (see Fig.~\ref{fig:siamese}). A Siamese-triplet network consists of three base networks with shared parameters, to process $R_1$, $R_2$ and $R^{-}$ in parallel. Based on the distance inequality we desire (discussed above), we use the loss function for triplets to enforce that region proposals that are not spatio-temporally close are further away than spatio-temporally close ones in the feature space. The idea of using triplets for learning embeddings has renewed interest lately~\cite{wang2014learning,schroff2015facenet,song2015deep} and can be traced to \cite{frome2007learning}, which uses triplets to learn an image-to-image distance function that satisfies the property that the distance between images from the same category should be less than the distance between images from different categories.

Specifically, given a triplet ($R_1$, $R_2$, $R^{-}$), where $R_1$ and $R_2$ are two spatio-temporally close region proposals and $R^{-}$ is a random region proposal from another video, the loss function in the feature space is defined as follows:
\vspace{-3pt}
	$$L(R_1,R_2,R^{-}) = \text{max}\{0, D(R_1, R_2) - D(R_1, R^{-}) + M\},$$ where $D(R_1, R_2)$ is the cosine distance of region proposals $R_1$ and $R_2$ in the feature space, and $M$ represents the margin between the two distances. This hinge loss forces the CNN to learn feature representations such that $D(R_1, R_2) < D(R_1, R^{-})-M$. In other words, the distance between the embeddings of spatio-temporally adjacent region proposals should be smaller than that between a query region proposal and a random region proposal, by a margin $M$. The overall objective function for training is as follows: 
\vspace{-5pt}
	$$\min_{W} \frac{\lambda}{2} ||W||_2^2 + \sum_{i=1}^{N} \max\{0,D(R_{i_1}, R_{i_2}) - D(R_{i_1}, R_i{-}) + M\},$$
\vspace{-1pt} 
where $W$ contains the parameter weights of the network, $N$ is the number of triplet samples and, $\lambda$ denotes the weight decay.

In our experiments, the convolutional layers of the base network follow the AlexNet architecture \cite{krizhevsky2012imagenet}, and two fully connected layers are stacked on the pool5 outputs, whose neuron numbers are 4096 and 1024 respectively. Mini-batch Stochastic Gradient Descent (SGD) is used during training. Initially, for each region proposal pair $(R_1, R_2)$, $K$ negative region proposal patches $R^{-}$ are randomly sampled in the same batch to form $K$ triplets. After training for several epochs, hard negative mining is used to enhance training. More specifically, for each pair $(R_1, R_2)$, the ranking losses of all other negative region proposals in the same batch are calculated, and the top $K$ ones with highest losses are selected to form $K$ hard triplets of samples. The idea is analogous to the hard-negative mining procedure in SVM. For details, please refer to \cite{wang2015unsupervised}.

\vspace{-12pt}

\subsection{Transferring Learnt Features for Supervised Tasks}

\label{fine-tune}
Until this point, our CNN has been trained in a purely unsupervised manner, using only unlabeled videos from YouTube. We evaluate how well features directly extracted from our unsupervised pre-trained models can benefit recognition tasks in Sec.~\ref{sec:unsupr}. To evaluate whether these features are useful for generic supervised recognition tasks, we can optionally adapt and specialize these purely unsupervised visual representations to tasks with labeled data. In our experiments in Sec.~\ref{sec:supr}, we fine-tune our pre-trained model on the PASCAL VOC multi-label classification task (VOC 2007 and VOC 2012) and MIT Indoor Scene single-label classification task.

We directly adapt our ranking model as a pre-trained network for the classification task. The adaptation method we use is similar to the approach applied in RCNN \cite{girshick2014rich}. However, RCNN uses the network pre-trained with ImageNet classification data (with semantic labels) as initialization of their supervised task. In our case, the pre-trained network is the \emph{unsupervised} CNN trained using unlabeled videos. More specifically, we use the weights of the convolutional layers in the base network of our Siamese-triplet architecture to initialize corresponding layers for the classification task. For weights of the fully connected layers, we initialize them randomly.

\vspace{-3pt}
\section{Experiments}

We present results on three datasets, with comparisons to several existing unsupervised methods~\cite{agrawal2015learning,wang2015unsupervised,pathak2016context} plus multiple informative baselines.  We consider both the purely unsupervised case where all learning stems from unlabeled videos (Sec.~\ref{sec:qual} and~\ref{sec:unsupr}) as well as the fine-tuning case where our method initializes a network trained with relatively few labeled images (Sec.~\ref{sec:supr}).

\label{training_procedures}
\vspace{-6pt}
\paragraph{Implementation details}

We use 25,000 unlabeled videos from YouTube downloaded from the first 25,000 URLs provided by Liang et al. \cite{liang2015towards}, which used thousands of keywords based on VOC to query YouTube. These unlabeled videos are of various categories, including movie trailers, animal/human activities, etc. Most video clips are dedicated to several objects, while some can contain hundreds of objects. We do not use any label information associated with each video. We extract video frames at the frame rate of 1 fps. Using Selective Search and our filtering process, we can easily obtain millions of region proposal pairs from these unlabeled videos. For efficiency, evaluation is throughout based on training our model with 1M region proposal pairs, which requires about 1 week to train. We can expect even better results if more data is used, though for greater computational expense.

We closely follow the Siamese-triplet network implementation from \cite{wang2015unsupervised} to learn our feature space. We set the margin parameter $M = 0.5$, weight decay $\lambda = 0.0005$, number of triplets per region proposal pair $K = 4$, and the batch size to be 100 in all experiments. The training is completed with Caffe \cite{jia2014caffe} based on 1M region proposal pairs (2M region proposal patches). We first train our model without hard negative mining at a constant learning rate $\epsilon = 0.001$ for 150K iterations. Then we apply hard negative mining with hard ratio 0.5 to continue training with initial learning rate $\epsilon_0 = 0.001$. We reduce the learning rate by a factor of 10 at every 100K iterations and train for another 300K iterations.
	
\vspace{-6pt}
\paragraph{Baselines}
We compare to several existing unsupervised feature learning methods \cite{agrawal2015learning,jayaraman2015learning,wang2015unsupervised,pathak2016context}. In all cases, we use the authors' publicly available pre-trained models. To most directly analyze the benefits of learning from region proposals, we implement three other baselines: \emph{full-frame}, \emph{square-region}, and \emph{visual-tracking}. For fair comparison, we use 1M frame/patch pairs as training data for each of these three baselines.  Note that all compared methods use unlabeled videos for pre-training, except for \cite{pathak2016context}, which uses unlabeled images. 
\subsubsection{\emph{Full-Frame}}
In every video, we take every two temporally adjacent video frames (1s apart) to be a positive pair. Namely, for a full-frame triplet ($R_1$, $R_2$, $R^{-}$), $R_1$ and $R_2$ are two temporally adjacent video frames and $R^{-}$ is a random full frame from another video. We take 1M full-frame pairs (2M full frames) as training data and follow the same procedures to train the baseline model.
\vspace{-15pt}
\subsubsection{\emph{Square-Region}}
Same as our proposed framework, we only generate square regions on selected video frame pairs using the proposed initial filtering process. For a selected temporally adjacent video frame pair (1s apart), we generate a random 227 $\times$ 227 patch in one frame and get the patch at the same position in the other frame. These two patches form a positive pair.  We repeat this process 10 times and generate 10 square-region pairs for every video frame pair. Namely, for a square-region triplet ($R_1$, $R_2$, $R^{-}$), $R_1$ and $R_2$ are two 227 $\times$ 227 patches from two temporally adjacent video frames at the same position, and $R^{-}$ is a random patch of the same size from another different video. We expect these square-region patches to be less object-like compared to patches obtained using our framework, because they are random regions from the frame. But note that square-region pairs also benefit from the same well-defined pruning steps we propose for our method. We take 1M square-region pairs (2M patches) as training data and follow the same procedures to train the baseline model.
\vspace{-15pt}
\subsubsection{\emph{Visual-Tracking}}
In \cite{wang2015unsupervised}, Wang et al. extract patches with motion and track these patches to create training instances. Specifically, they first obtain SURF \cite{bay2006surf} interest points and use Improved Dense Trajectories (IDT) \cite{wang2013action} to obtain the motion of each SURF point. Then they find the best bounding box such that it contains most of the moving SURF points. After obtaining the initial bounding box (the first patch), they perform tracking using the KCF tracker \cite{henriques2015high} and track along 30 frames in the video to obtain the second patch. These two patches form a positive pair. We take 1M visual-tracking pairs (2M patches) from their publicly available dataset of collected patches as our training data and follow the same procedures to train the baseline model.
	\vspace{-15pt}
	\subsubsection{\emph{Random-Gaussian}}
As a sanity check, we also provide a random initialization baseline.  Specifically, we construct a CNN using the AlexNet architecture, and initialize all layers with weights drawn from a Gaussian distribution. This randomly initialized AlexNet serves as a pre-trained model.
	\vspace{-1em}

\vspace{-3pt}	
\subsection{Qualitative Results of Unsupervised Feature Learning}
\label{sec:qual}
\vspace{-3pt}	

\subsubsection{Visualization of Filters}

\begin{figure}
  \centering
    \begin{subfigure}[b]{0.23\textwidth}
        \includegraphics[width=\textwidth]{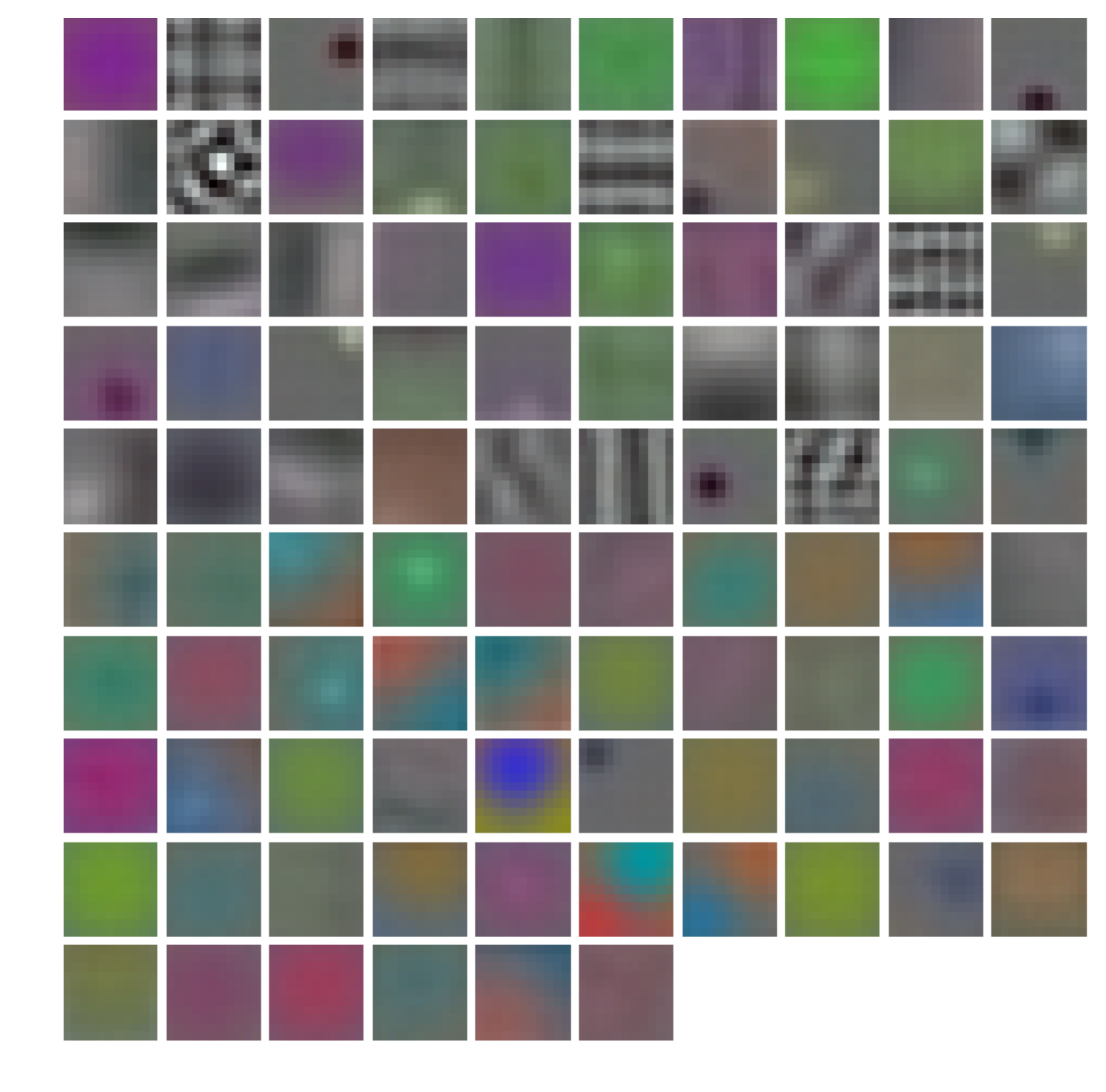}
        \caption{Ours}
        \label{fig:ous}
    \end{subfigure}
    \hfill
    ~ 
    \begin{subfigure}[b]{0.23\textwidth}
        \includegraphics[width=\textwidth]{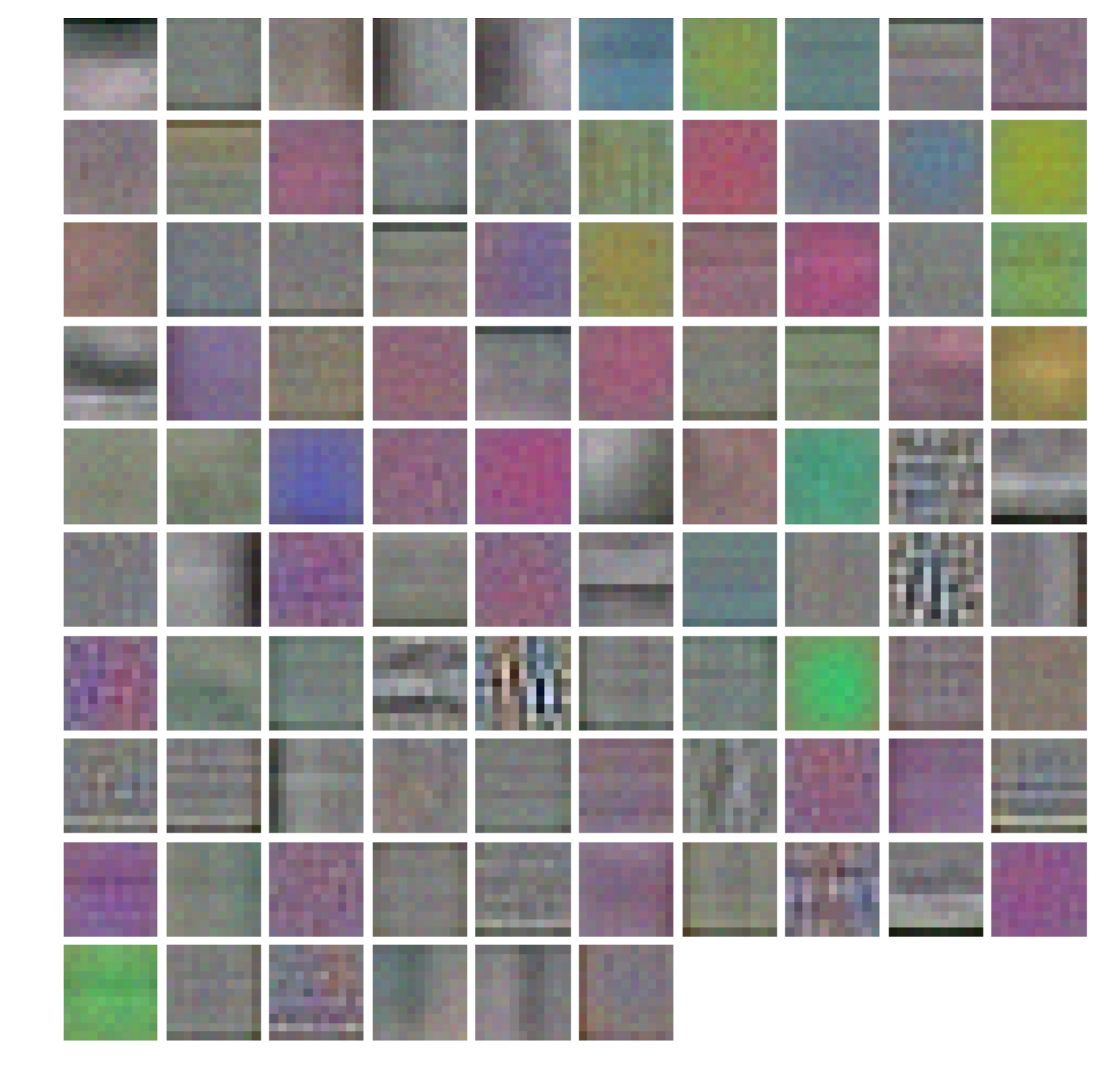}
        \caption{Full-Frame}
        \label{fig:full}
    \end{subfigure}
    \hfill
    ~ 
    \begin{subfigure}[b]{0.23\textwidth}
        \includegraphics[width=\textwidth]{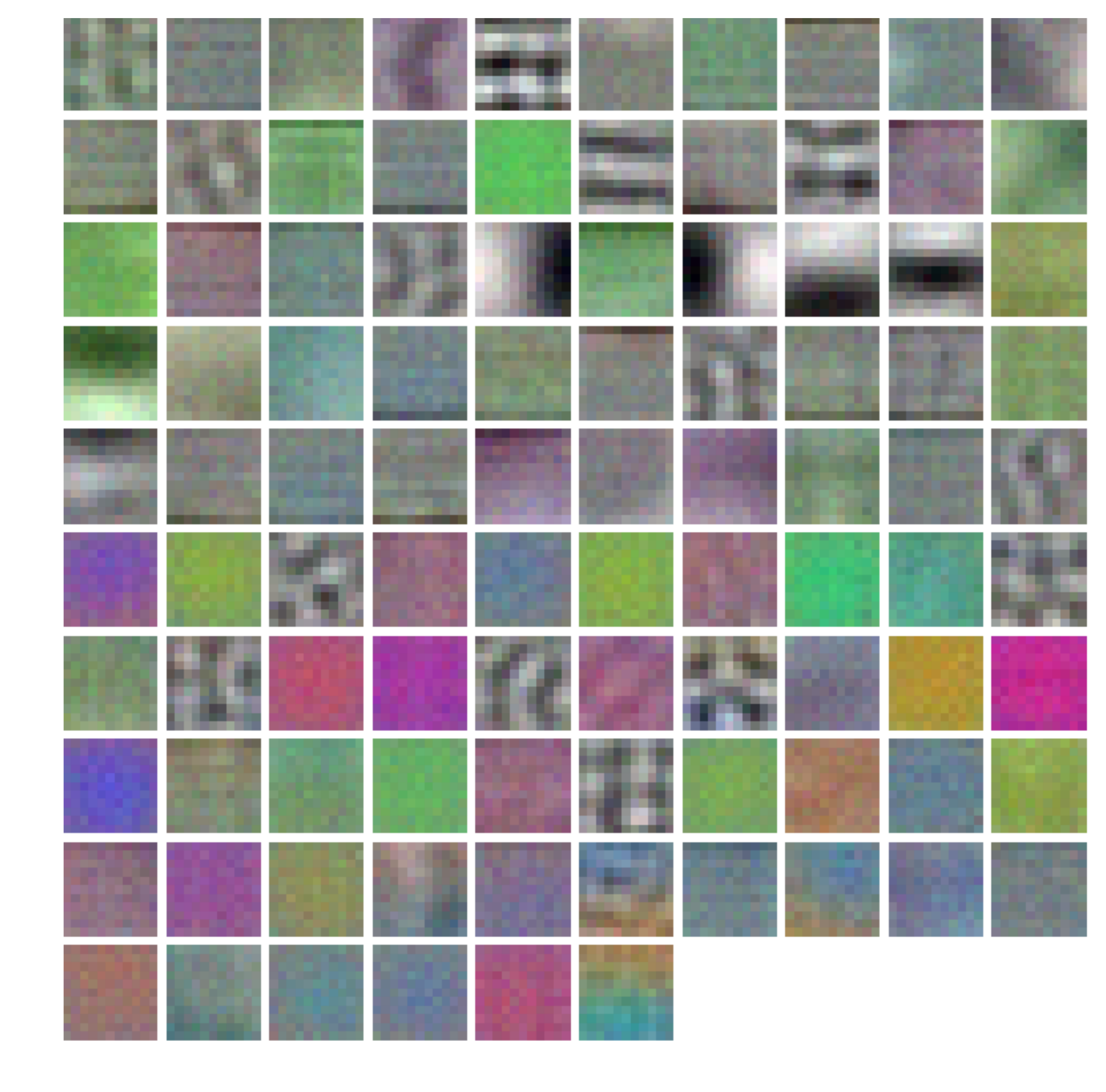}
        \caption{Square-Region}
        \label{fig:random}
    \end{subfigure}
    \hfill
    ~ 
    \begin{subfigure}[b]{0.23\textwidth}
        \includegraphics[width=\textwidth]{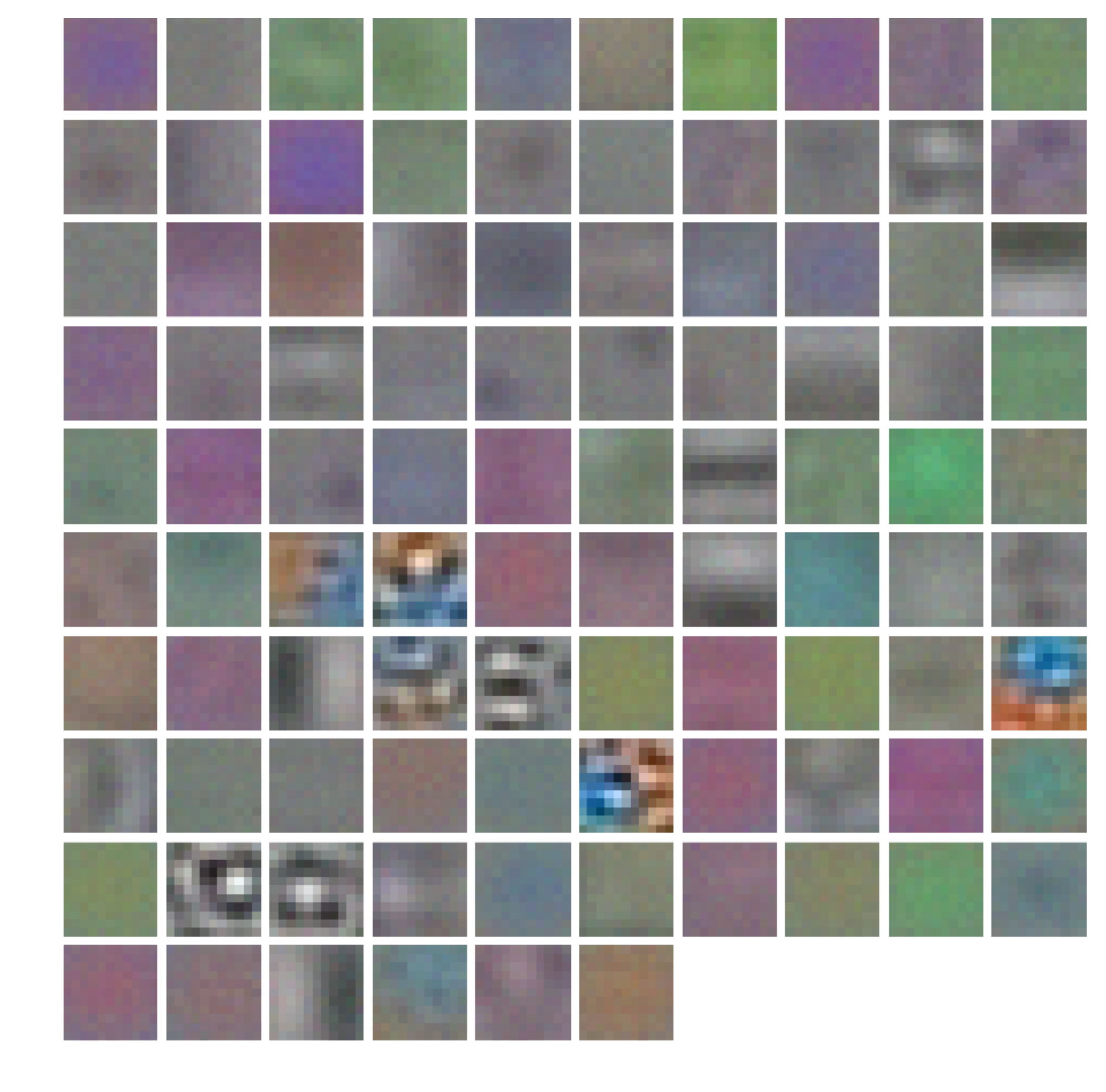}
        \caption{Visual-Tracking}
        \label{fig:motion}
    \end{subfigure}
    \hfill
    \caption{Visualization of Conv1 Features}\label{fig:conv1_features}
\end{figure}

We first analyze our learnt features qualitatively. We visualize the conv1 features learnt in our unsupervised CNN as well as the first three baselines above. The visualization is shown in Fig.~\ref{fig:conv1_features}. We observe that the conv1 features learnt using our framework are much more distinctive than the three baselines, suggesting a more powerful basis.

\begin{figure}
\vspace{-0.5em}
      \centering
      \vspace{-0.5em}
      \includegraphics[scale=0.40]{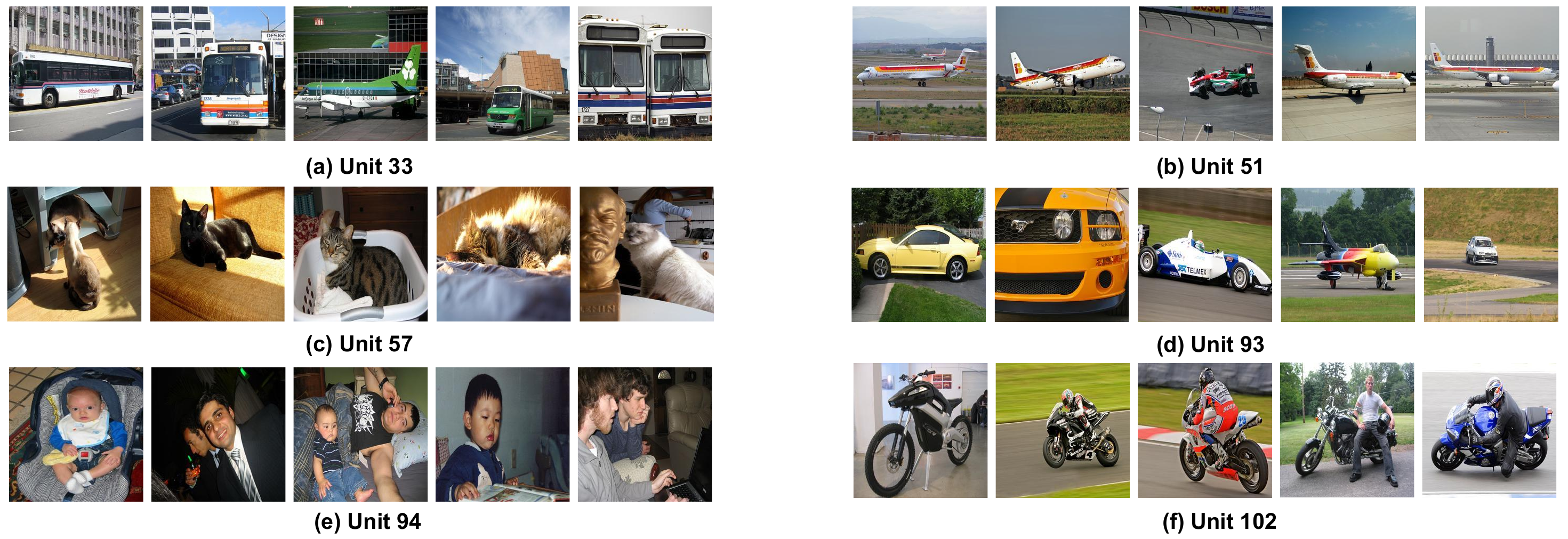}
            \caption{Top response images from PASCAL VOC 2012 for 6 of the conv5 neurons of our unsupervised CNN. Each seems to correspond to one semantically meaningful category, despite no labeled data during learning. }
            \vspace{-1em}
            \label{fig:activation}
\end{figure}

To better understand the internal representations of the units learnt by our unsupervised CNN, in Fig.~\ref{fig:activation}, we get the top response images for units in the conv5 layer. We use all images in PASCAL VOC 2012 as the database. Although here we do \emph{not} provide any semantic information during training, we can see that units in conv5 layers nonetheless often fire on semantically meaningful categories. For example, unit 33 fires on buses, unit 57 fires on cats, etc.

\vspace{-24pt}
\subsubsection{Neighbors in Unsupervised Learned Feature Space}

Next we analyze the learned features via a retrieval task.  We perform Nearest Neighbors (NN) using ground-truth (GT) windows in the PASCAL VOC 2012 val set as queries, and a retrieval database consisting of all selective search windows (having more than 0.5 overlap with GT windows) in the PASCAL VOC 2012 train set. 

\begin{figure}[t]
      \centering
      \includegraphics[scale=0.36]{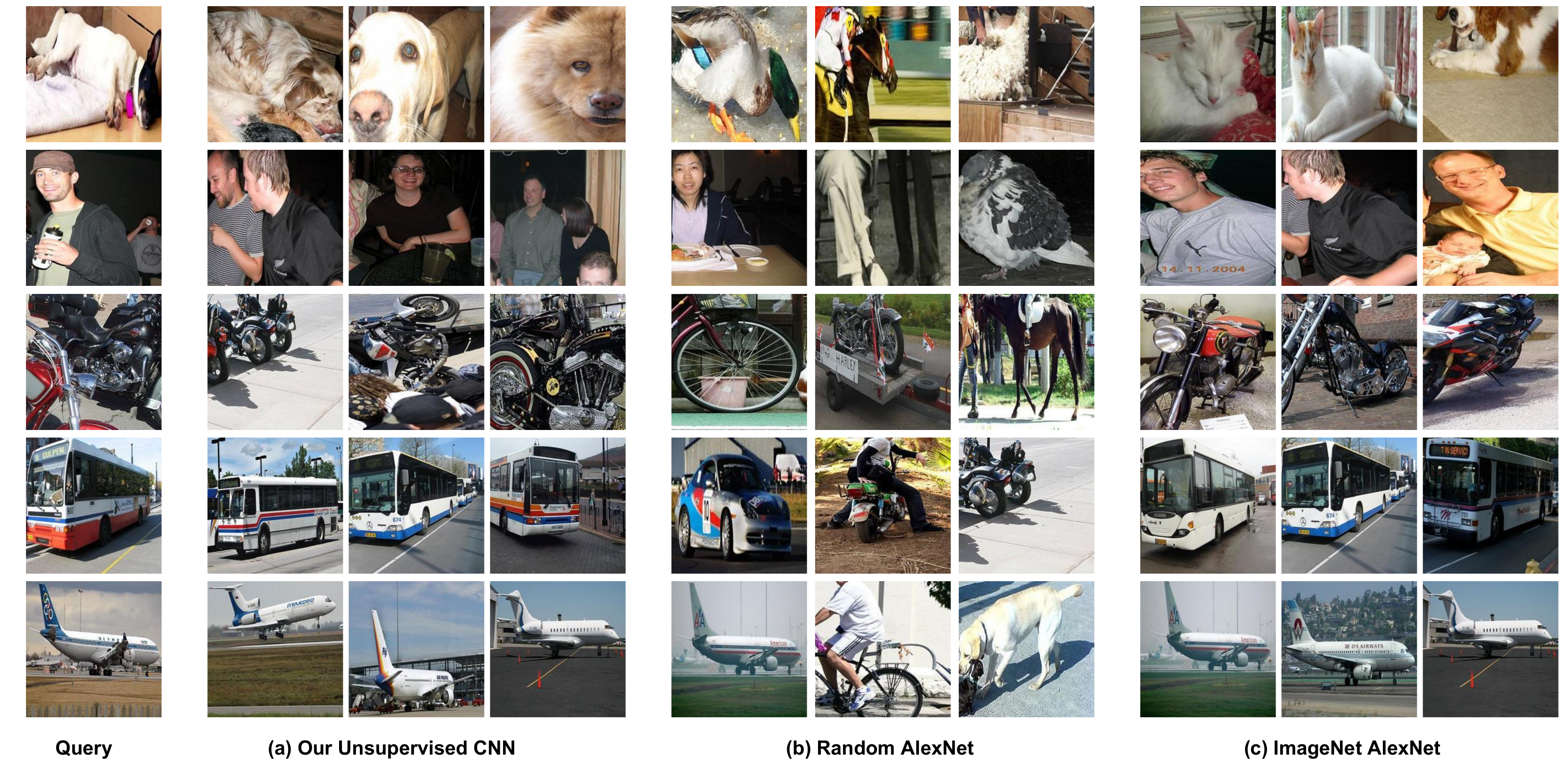}
            \caption{Nearest Neighbor Results. We compare three models: (a) Our unsupervised CNN pre-trained using 1M region proposal pairs; (b) AlexNet initialized with random parameters; (3) AlexNet trained with labeled ImageNet. Our unsupervised CNN is far superior to random AlexNet, and even learns some concepts similar to ImageNet AlexNet trained with semantic labels.}
            \label{fig:NN}
\end{figure}

Fig.~\ref{fig:NN} shows example results. We can see that our unsupervised CNN is far superior to AlexNet initialized with random weights.  Furthermore, despite having access to zero labeled data, our features (qualitatively) appear to produce neighbors comparable to AlexNet trained on ImageNet with over 1 million semantic labels. For example, in the first row, given an image of a dog lying down as a query, our method successfully retrieves images of dogs of different postures. It even outperforms ImageNet AlexNet, which retrieves two cats as nearest neighbors. In comparison, AlexNet initialized with random weights mostly retrieves unrelated images. Note that there is no class supervision or fine-tuning being used here in retrieval for our unsupervised CNN, and \emph{all the gains for our pre-trained model come from unlabeled videos.} 

Quantitatively, we can measure the retrieval rate by counting the number of correct retrievals in the top-20 neighbors. Given a query image, a retrieval is considered correct if the semantic class for the retrieved image and the query image are the same. Using pool5 features extracted from our unsupervised CNN and cosine distance, we obtain 32\% retrieval rate, which outperforms 17\% by random AlexNet (our method's initialization) by a large margin. ImageNet AlexNet achieves 62\% retrieval rate, but note that it is provided with substantial labeled data from which to directly learn semantics.

\vspace{-6pt}
\subsection{Unsupervised Feature Learning Recognition Results}
\vspace{-6pt}
\label{sec:unsupr}

Next we evaluate how well our unsupervised approach can benefit recognition tasks---\emph{without} any network fine-tuning with labeled data.  We test  on three datasets: MIT Indoor 67~\cite{quattoni2009recognizing}, PASCAL VOC 2007 and 2012.
 
For MIT Indoor 67, the database contains 67 indoor categories, and a total of 15,620 images. We use the subset defined in \cite{quattoni2009recognizing}. The training set contains 5360 (67$\times$80) images and the test set contains 1340 (67$\times$20) images. For PASCAL VOC 2007, we use all single-labeled images, namely 3103 single-labeled images from the trainval set as training data, and 3192 single-labeled images in the test set as testing data.  Similarly, for PASCAL VOC 2012, this amounts to 3730 single-labeled training images (train set) and 3837 test images (val set).  For all our baselines, we extract pool5 features from the network and train a linear classifier using softmax loss.
 
Table~\ref{table:unsupervised} shows the results. The third row shows the results of four unsupervised pre-trained models using other approaches (using their publicly available models), and the fourth row shows the results of our four baselines. Overall, our method obtains gains averaging 20\% over the existing methods, and 30\% over the additional baselines. 

Our unsupervised pre-trained model is much better than a randomly initialized AlexNet. Note that we do not use any label information or fine-tuning here; our results are obtained using our unsupervised CNN trained purely on unlabeled videos. We also out-perform pre-trained models from Pathak et al. \cite{pathak2016context}, Jayaraman et al. \cite{jayaraman2015learning} and Agrawal et al. \cite{agrawal2015learning} by quite a large margin (around 20\%). The pre-trained model from Wang et al. \cite{wang2015unsupervised} has better performance, but their model is trained using substantially more data (4M visual-tracking pairs compared to our 1M pairs). 

For fair and efficient comparison with our method and baselines, we take 1M of the collected visual-tracking pairs from~\cite{wang2015unsupervised} and implement their method. This is our ``visual-tracking'' baseline. Our approach outperforms all our baselines, including ``visual tracking'' on all three datasets. Surprisingly, our square-region baseline also has impressive performance. We attribute its competitive performance to our well-defined filtering process. Although the patches used for square-region baseline may not correspond to a certain object or object part, the two patches that form the square-region pair are guaranteed to be relevant. Moreover, square-region patches are taken randomly from the whole frame, and therefore it has no bias towards either objects or scenes.

\vspace{-1em}
\begin{table}
\centering
\begin{tabular}{c|c|c|c|c}
\specialrule{1pt}{1pt}{1pt}
Method    & Supervision                     & MIT Indoor 67 & VOC 2007 & VOC 2012\\ \hline
ImageNet              & 1.2M labeled images               & 54\%  & 71\%  & 72\%                 \\  \hline
Wang et al. \cite{wang2015unsupervised}          & 4M visual tracking pairs                          & 38\%     & 47\%      & 48\%    \\
Jayaraman et al. \cite{jayaraman2015learning}       & egomotion                         & 26\%   & 40\%     & 39\%      \\
Agrawal et al. \cite{agrawal2015learning}       & egomotion                       & 25\%   & 38\%     & 37\%       \\
Pathak et al. \cite{pathak2016context}       & spatial context                         & 23\%   & 36\%     & 36\%       \\ \hline
Full-Frame            & 1M video frame pairs  & 27\%  & 40\%  & 40\%             \\
Square-Region         & 1M square region pairs & 32\%  & 42\%  & 42\%               \\
Visual-Tracking \cite{wang2015unsupervised}       & 1M visual tracking pairs & 31\% & 42\%    & 42\%              \\
Random Gaussian & -                               & 16\%  & 30\%    & 28\%
	\\ \hline
Ours                  & 1M region proposal pairs       &         34\% & 46\% & 47\%     \\
\specialrule{1pt}{1pt}{1pt}
\end{tabular}
\vspace{0.5em}
\caption{Quantitative comparisons for image classification on MIT Indoor 67, PASCAL VOC 2007 and PASCAL VOC 2012. The third outlined row shows the results of four unsupervised pre-trained models using other approaches. The fourth outlined row shows the results of our four baselines. The visual-tracking baseline is the same approach as Wang et al. \cite{wang2015unsupervised}, but uses the same amount of data as ours to train the model for fairest comparison.}
\label{table:unsupervised}
\vspace{-3.5em}
\end{table}

\vspace{-1em}
\subsection{Fine-tuned Feature Learning Recognition Results}
\label{sec:supr}

Finally, we evaluate our learnt features on image classification tasks after fine-tuning on the three datasets. For PASCAL VOC 2007, we fine-tune using VOC 2007 trainval set (5011 images) and test on VOC 2007 test set (4952 images). For PASCAL VOC 2012, we fine-tune using VOC 2012 train set (5717 images) and test on VOC 2012 val set (5823 images). For MIT Indoor 67, we fine-tune using the training set (5360 images) and test on the test set (1340 images). We use a simple horizontal flip as data augmentation and fine-tune for the same number of iterations for all methods. For PASCAL VOC multi-label classification tasks, we use the standard mean Average Precision (mAP) to evaluate the predictions. For MIT Indoor Scene single-label classification task, we report the classification accuracy.

Table~\ref{table:pascal_classification} shows the results. Note that for the fine-tuning task, most of the learning for the network comes from labeled images in the fine-tuning dataset. The pre-trained models serve only as an initialization for fine-tuning. Our method consistently outperforms all our baseline methods and the pre-trained models from Pathak et al. \cite{pathak2016context}, Jayaraman et al. \cite{jayaraman2015learning} and Agrawal et al. \cite{agrawal2015learning}.  The pre-trained model from \cite{wang2015unsupervised} has better performance if using four times the training data as our method. However, for the comparable setting with the same amount of input video and identical fine-tuning procedures, our method is superior to the tracking-based approach (see Visual-Tracking~\cite{wang2015unsupervised} row). This comparison is the most direct and speaks favorably for the core proposed idea.
  
\vspace{-1.5em}
\begin{table}
\centering
\begin{tabular}{c|c|c|c|c}
\specialrule{1pt}{1pt}{1pt}
Pretraining Method    & Supervision         & MIT Indoor 67           & VOC 2007 & VOC 2012\\ \hline
ImageNet              & 1.2M labeled images       & 61.6\%        & 71.1\%  & 70.2\%                         \\ \hline
Wang et al. \cite{wang2015unsupervised}          & 4M visual tracking pairs   & 41.6\%                       & 47.8\%     & 47.4\%          \\
Jayaraman et al. \cite{jayaraman2015learning}       & egomotion       & 31.9\%                  & 41.7\%   & 40.7\%            \\ 
Agrawal et al. \cite{agrawal2015learning}       & egomotion           & 32.7\%            & 42.4\%   & 40.2\%            \\
Pathak et al. \cite{pathak2016context}       & spatial context        & 34.2\%                 & 42.7\%   & 41.4\%            \\  \hline
Full-Frame            & 1M video frame pairs  & 33.4\% & 41.9\% & 40.3\%               \\
Square-Region         & 1M square region pairs  & 35.4\% & 43.2\% & 42.3\%              \\
Visual-Tracking \cite{wang2015unsupervised}      & 1M visual tracking pairs & 36.6\% & 43.6\%   & 42.1\%            \\
Random Gaussian & -                            & 28.9\%  & 41.3\%  & 39.1\%
	\\ \hline
Ours                  & 1M region proposal pairs  &  38.1\% & 45.6\% & 44.1\%    \\
\specialrule{1pt}{1pt}{1pt}
\end{tabular}
\vspace{0.5em}
\caption{Classification results on MIT Indoor 67, PASCAL VOC 2007 and PASCAL VOC 2012. Accuracy is used for MIT Indoor 67 and mean average precision (mAP) is used for PASCAL VOC to compare the models. The third row shows the results of four unsupervised pre-trained models using other approaches. The fourth row shows the results of our four baselines.}
\label{table:pascal_classification}
\vspace{-3em}
\end{table}
\vspace{-3pt}
\section{Conclusion and Future Work}
\vspace{-3pt}
We proposed a framework to learn visual representations from unlabeled videos. Our approach exploits object-like region proposals to generate associated regions across video frames, yielding localized patches for training an invariant embedding without explicit tracking.  Through various experiments on image retrieval and image classification, we have shown that our method provides useful feature representations despite the absence of strong supervision.  Our method outperforms multiple existing approaches for unsupervised pre-training, providing a new promising tool for the feature learning toolbox.   In future work we plan to consider how spatio-temporal video object segmentation methods could enhance the proposals employed to create training data for our framework.
\\\\
\textbf{Acknowledgements:} This research is supported in part by ONR PECASE N00014-15-1-2291. We also thank Texas Advanced Computing Center for their generous support and the anonymous reviewers for their comments.
\bibliographystyle{splncs}
\bibliography{cv_archive.bib}
\end{document}